\pdfoutput=1
\documentclass[11pt]{article}
\usepackage{emnlp2021}
\usepackage{times}
\usepackage{latexsym}
\usepackage[T1]{fontenc}
\usepackage[utf8]{inputenc}
\usepackage{microtype}
\usepackage{times}
\usepackage{latexsym}
\usepackage{url}
\usepackage{booktabs}
\usepackage{appendix}
\usepackage{xcolor}
\usepackage{algorithm}
\usepackage{algorithmic}
\usepackage{amsmath}
\usepackage{amsfonts}
\usepackage{natbib}
\usepackage{multirow}
\usepackage{tikz}
\usepackage{graphicx}
\usepackage{caption, subcaption}

\title{Identifying Human Strategies for Generating Word-Level Adversarial Examples}

\author{
 Maximilian Mozes$^1$\quad Bennett Kleinberg$^{1,2}$\quad Lewis D. Griffin$^1$\\
 $^1$University College London\\
 $^2$Tilburg University\\
 \small{\texttt{\{m.mozes, l.griffin\}@cs.ucl.ac.uk}} \\
 \small{\texttt{bennett.kleinberg@tilburguniversity.edu}}
}

\begin{document}
\maketitle
\begin{abstract}
Adversarial examples in NLP are receiving increasing research attention. One line of investigation is the generation of word-level adversarial examples against fine-tuned Transformer models that preserve naturalness and grammaticality. Previous work found that human- and machine-generated adversarial examples are comparable in their naturalness and grammatical correctness. Most notably, humans were able to generate adversarial examples much more effortlessly than automated attacks. In this paper, we provide a detailed analysis of exactly how humans create these adversarial examples. By exploring the behavioural patterns of human workers during the generation process, we identify statistically significant tendencies based on which words humans prefer to select for adversarial replacement (e.g., word frequencies, word saliencies, sentiment) as well as \textit{where} and \textit{when} words are replaced in an input sequence. With our findings, we seek to inspire efforts that harness human strategies for more robust NLP models.
\end{abstract}

\section{Adversarial attacks in NLP}
Researchers in natural language processing (NLP) have identified the vulnerability of machine learning models to adversarial attacks: controlled, meaning-preserving input perturbations that cause a wrong model prediction~\cite{jia-liang-2017-adversarial,iyyer-etal-2018-adversarial,ribeiro-etal-2018-semantically}. 
Such adversarial examples uncover model failure cases and are a major challenge for trustworthiness and reliability. While several defence methods exist against adversarial attacks~\cite{huang-etal-2019-achieving,jia-etal-2019-certified,zhou-etal-2019-learning,jones-etal-2020-robust,le-etal-2022-shield}, developing robust NLP models is an open research challenge. An in-depth analysis of word-level adversarial examples, however, identified a range of problems, showing that they are often ungrammatical or semantically inconsistent~\cite{morris-etal-2020-reevaluating}.\footnote{For example, replacing the word \textit{summer} with \textit{winter}.} This finding raised the question of how feasible natural and grammatically correct adversarial examples actually are in NLP.

To answer this question,~\citet{mozes2021contrasting} explored whether humans are able to generate adversarial examples that are valid under such strict requirements. In that study, crowdworkers were tasked with the generation of word-level adversarial examples against a target model. The findings showed that at first sight---without strict validation---humans are less successful than automated attacks. However, when adding constraints on the preservation of sentiment, grammaticality and naturalness, human-authored examples do not differ from automated ones. The most striking finding was that automated attacks required massive computational effort while humans were able to do the same task using only a handful of queries.\footnote{For example, 140,000 queries are needed per example for \textsc{SememePSO}~\cite{zang-etal-2020-word}, on average, to generate successful adversarial examples on IMDb~\cite{Maas2011}, whereas humans need 10.9 queries~\cite{mozes2021contrasting}.} This suggests that humans are far more efficient in adversarial attacks than automated systems, yet exactly how they achieve this is unclear.

In this work, we address this question by analysing human behaviour through the public dataset from ~\citet{mozes2021contrasting}. We look at which words humans perturbed, where within a sentence those perturbations were located, and whether they mainly focused on perturbing sentiment-loaded words. We find that \textit{(i)} in contrast to automated attacks, humans use more frequent adversarial word substitutions, \textit{(ii)} the semantic similarity between replaced words and adversarial substitutions is greater for humans than for most attacks, and \textit{(iii)} humans replace sentiment-loaded words more often than algorithmic attackers. Our goal is to understand what makes humans so efficient at this task, and whether these strategies could be harnessed for more adversarially robust NLP models.

\section{Data and Models}
We present a fine-grained analysis of the strategies that human crowdworkers employed to generate word-level adversarial examples against sentiment classification models. In the dataset from ~\citet{mozes2021contrasting}, 43 participants were recruited via Amazon Mechanical Turk and trained to perform a word-level adversarial attack on test set sequences from the IMDb movie reviews dataset~\cite{Maas2011}. In total, 170 adversarial examples were collected. For each of the collected adversarial examples, the authors also generated automated adversarial examples using the \textsc{TextFooler}~\cite{jin2019bert}, \textsc{BAE}~\cite{garg-ramakrishnan-2020-bae}, \textsc{Genetic}~\cite{alzantot-etal-2018-generating} and \textsc{SememePSO}~\cite{zang-etal-2020-word} attacks. 

The \textsc{TextFooler} attack uses a greedy word-replacement algorithm that is guided by word saliencies and semantic similarity measures between an unperturbed sequence and the adversarial candidate.
The \textsc{BAE} attack resorts to a different technique, utilising a BERT-based language model to remove and replace tokens in an input sequence.
The \textsc{Genetic} attack, in contrast, is based on a population-based method using genetic algorithms. Finally, the \textsc{SememePSO} attack is based on replacements of word sememes instead of entire words and combines this with a particle swarm optimisation approach.

All attacks were performed against a RoBERTa model~\cite{liu2019roberta} fine-tuned on IMDb.\footnote{For more model details, see Section 3 in~\citet{mozes2021contrasting}.} Here, we only consider adversarial examples that preserved sentiment after evaluation by an independent set of crowdworkers, which \citet{mozes2021contrasting} used as a key validity criterion. 

\begin{table}[!t]
\resizebox{1.0\columnwidth}{!}{
	\centering
		\begin{tabular}{l c c c c c c c c c}
			\toprule
		    \multicolumn{1}{c}{\multirow{2}[2]{*}{\textbf{Attack}}} & 
		    \multicolumn{3}{c}{\textbf{All}} & 
		    \multicolumn{3}{c}{\textbf{Successful}} &
		    \multicolumn{3}{c}{\textbf{Unsuccessful}} \\
            \cmidrule(lr){2-4}\cmidrule(lr){5-7}\cmidrule(lr){8-10}
            & $\Delta_{M}$ & $\Delta_{SD}$ & $d$ & $\Delta_{M}$ & $\Delta_{SD}$ & $d$ & $\Delta_{M}$ & $\Delta_{SD}$ & $d$ \\
			\midrule
			\textsc{HumanAdv} & 0.6 & 3.1 & 0.2 & 0.5 & 3.0 & 0.1 & 0.6 & 3.1 & 0.2 \\
			\textsc{TextFooler} & 2.5 & 2.6 & 0.8 & 2.5 & 2.6 & 0.8 & 2.5 & 2.6 & 0.8 \\
			\textsc{Genetic} & 1.5 & 2.1 & 0.5 & 1.4 & 2.0 & 0.5 & 1.5 & 2.1 & 0.5 \\
			\textsc{BAE} & 2.0 & 4.0 & 0.5 & 1.9 & 4.1 & 0.5 & 2.0 & 4.0 & 0.5 \\
			\textsc{SememePSO} & 2.4 & 2.8 & 0.8 & 2.4 & 2.8 & 0.8 & -- & -- & -- \\
			\bottomrule
		\end{tabular}
	}
	\caption{Word frequency differences between replaced words and adversarial substitutions. $\Delta_{M}$ and $\Delta_{SD}$ represent the mean and standard deviation of the differences between replaced words and substitutions (i.e., positive values: replaced words $>$ substitutions), $d$ denotes the Cohen's $d$ effect size. Note that for \textsc{SememePSO}, all adversarial examples are successful.}
	\label{tab:freq_diffs}
\end{table}

\section{Analysis}
\label{sec:analysis}
In this section, we report on a series of experiments analysing the human- and machine-authored adversarial examples. 

\subsection{What do humans replace?}

\paragraph{Word frequency.}
We investigate the word frequency of the adversarial examples. Existing work~\cite{mozes2021frequency, hauser2021bert} identified significant differences in word frequency between adversarially perturbed words (hereafter referred to as \textit{replaced words}) and their substitutions (hereafter referred to as \textit{adversarial substitutions}) for a number of attacks. The substituted words were considerably less frequent than their original counterparts (e.g., \textit{annoying} $\to$ \textit{galling}).\footnote{Word frequency is computed with respect to the model's training corpus in these experiments.} 
Here, we examine whether this pattern is also evident in humans' strategies. Table~\ref{tab:freq_diffs} shows the differences of the $\log_e$ word frequencies between replaced words and corresponding substitutions for all four automated adversarial attacks and the human attack. All attacks replace words with less frequent substitutions. The notable observations here are the human-authored examples: the $\log_e$ frequency differences are lowest for the human-generated substitutions (\textsc{HumanAdv}). 
The effect size Cohen's $d$, which expresses the absolute magnitude of the effect that frequencies differ, further shows that the high-to-low frequency replacement is much less used by humans ($d=0.2$) than by the other, automated attacks ($d \geq 0.5$). These findings persist when inspecting either successful or unsuccessful adversarial examples in isolation.

To test for statistical differences between the attacks, we first conduct a 5 (attacks) by 2 (success) ANOVA on the $\log_e$ frequency differences between replaced words and substitutions, to determine whether main effects or interaction effects were present. We observe a significant main effect for attack, $F(4, 12003)=152.85, p<.001$, but none for success nor an interaction between attack and success.\footnote{Follow-up experiments revealed significant differences between \textsc{HumanAdv} and all attacks.}

Overall, the results suggest that humans use a strategy different from automated approaches and find replacements that do not rely on the high-to-low frequency mapping to the same extent as automated attacks. Illustrations of the highest and lowest frequency differences among word substitution pairs can be found in the Appendix (Table~\ref{tab:words_freq_diffs}).

\paragraph{Word saliency usage.}
In the crowdsourcing study by ~\citet{mozes2021contrasting}, humans were provided with the word saliency information (i.e., individual words were highlighted based on how much the model's prediction confidence would change if they were deleted).\footnote{It is worth noting that we cannot be certain whether humans did indeed use the word saliencies during the process.} This was originally intended to make the task easier for humans. Now, we investigate whether humans did indeed focus on salient words.

\paragraph{Did humans prefer salient words?}
First we investigate whether the saliency of a replaced word correlates with the iteration index at which this word was selected for replacement by a human crowdworker.\footnote{An iteration index of 1 means that a word was the first to be replaced.} Across all examples, we obtain a negligible negative Pearson correlation of $r=-0.05$ ($p<0.01$). 
However, the correlation is weak, which does not provide additional evidence  in favour of utilising saliencies for automated attacks based on human behaviour.

\paragraph{Did salient words lead to successful attacks?}
We furthermore analyse whether the average saliency across all replaced words of a sequence correlates with whether this led to a successful (i.e., label-flipping) adversarial example. For each valid human-generated adversarial example, we hence compute the point-biserial correlation between attack success and mean saliency of replaced words. The findings suggest that the higher the saliency of replaced words, the higher the chance of success of an adversarial example, $r=0.26$ ($p < .006$). Analogously, we also computed the correlation between the mean word saliency across all replaced words per iteration and the corresponding decrease in prediction confidence. The findings indicate a small correlation of $r=0.12$ ($p<.001$): replacing a more salient word leads to larger increases in prediction confidence change. 

It is worth noting that, even though the word saliency is defined as the decrease in prediction confidence after deleting a word from the sequence, this finding is not necessarily expected: a human attacker not only needs to identify and remove an existing word in the sequence, but they also have to find a semantically suitable replacement that decreases the model’s prediction confidence.

It is furthermore worth mentioning that both \textsc{BAE} and \textsc{TextFooler} define the token importance rankings based on a word saliency measure, and therefore explicitly incorporate the word saliency into the attack process. The results obtained in this work provide additional evidence in favour of utilising saliencies for automated attacks, showing that humans (which have been shown to generate adversarial examples in a more efficient way) also tend to utilise word saliencies.

\paragraph{Word similarities.}
Next, we compare the semantic differences in adversarial substitution pairs across the different attacks. While the algorithmic attacks source word synonyms from available lexical databases such as \textsc{WordNet}~\cite{fellbaum98wordnet} or \textsc{GloVe} embeddings~\cite{pennington-etal-2014-glove}, humans directly choose word replacements based on their own vocabulary and can therefore use substitutions that more accurately fit the context of the replaced word. Hence, we might expect to see a difference between the semantic similarity of human- and machine-chosen substitutions.

\begin{table}[!t]
    \resizebox{1.0\columnwidth}{!}{
    	\centering
    		\begin{tabular}{l c c c c }
    			\toprule
    		    \textbf{Attack} & 
    		    \textbf{Valid pairs} & 
    		    \textbf{All} & 
    		    \textbf{Succ.} &
    		    \textbf{Unsucc.} \\
    			\midrule
    			\textsc{HumanAdv} & 990/1303 & 0.47 (0.19) & 0.52 (0.18) & 0.44 (0.19) \\
    			\textsc{TextFooler}$^{a,b}$ & 1497/1805 & 0.57 (0.20) & 0.58 (0.20) & 0.53 (0.16) \\
    			\textsc{Genetic}$^{b}$ & 1955/2437 & 0.44 (0.19) & 0.44 (0.18) & 0.44 (0.19) \\
    			\textsc{BAE}$^{a,b}$ & 940/1623 & 0.69 (0.25) & 0.70 (0.24) & 0.69 (0.26) \\
    			\textsc{SememePSO}$^{b}$ & 724/946 & 0.66 (0.17) & 0.66 (0.17) & -- \\
    			\bottomrule
    		\end{tabular}
	}
	\caption{The mean (SD) cosine distances between replaced words and substitutions. $^{a}$ indicates significant differences with \textsc{HumanAdv} for unsuccessful pairs, $^{b}$ for successful ones.}
	\label{tab:emb_dists}
\end{table}

\begin{table}[!t]
    \resizebox{1.0\columnwidth}{!}{
    	\centering
    		\begin{tabular}{l c c c c c c}
    			\toprule
    		    \multicolumn{1}{c}{\multirow{2}[2]{*}{\textbf{Attack}}} & 
    		    \multicolumn{2}{c}{\textbf{All}} & 
    		    \multicolumn{2}{c}{\textbf{Succ.}} &
    		    \multicolumn{2}{c}{\textbf{Unsucc.}} \\
                \cmidrule(lr){2-3}\cmidrule(lr){4-5}\cmidrule(lr){6-7} 
                & \textbf{Rep.} & \textbf{Sub.} & \textbf{Rep.} & \textbf{Sub.} & \textbf{Rep.} & \textbf{Sub.} \\
    			\midrule
    			\textsc{HumanAdv} & 22.9 & 20.7 & 23.7 & 24.0 & 22.5 & 19.3 \\
    			\textsc{TextFooler}$^b$ & 19.8 & 14.2 & 19.8 & 14.3 & 18.8 & 12.5 \\
    			\textsc{Genetic}$^b$ & 19.7 & 14.3 & 20.3 & 15.7 & 19.6 & 14.0 \\
    			\textsc{BAE}$^{a,b}$ & 16.5 & 4.3 & 19.3 & 5.3 & 15.8 & 4.0 \\
    			\textsc{SememePSO} & 21.8 & 20.8 & 21.8 & 20.8 & -- & -- \\
    			\bottomrule
    		\end{tabular}
	}
	\caption{Ratio (\%) of replaced (Rep.) and adversarially substituted words (Sub.) with existing sentiment value. $^a$ indicates significant differences with \textsc{HumanAdv} for replaced words, $^b$ for substitutions.}
	\label{tab:sent_values}
\end{table}

\begin{figure*}[t]
\resizebox{1.0\textwidth}{!}{
    \subfloat[\textsc{HumanAdv}]{{\includegraphics[width=0.50\textwidth]{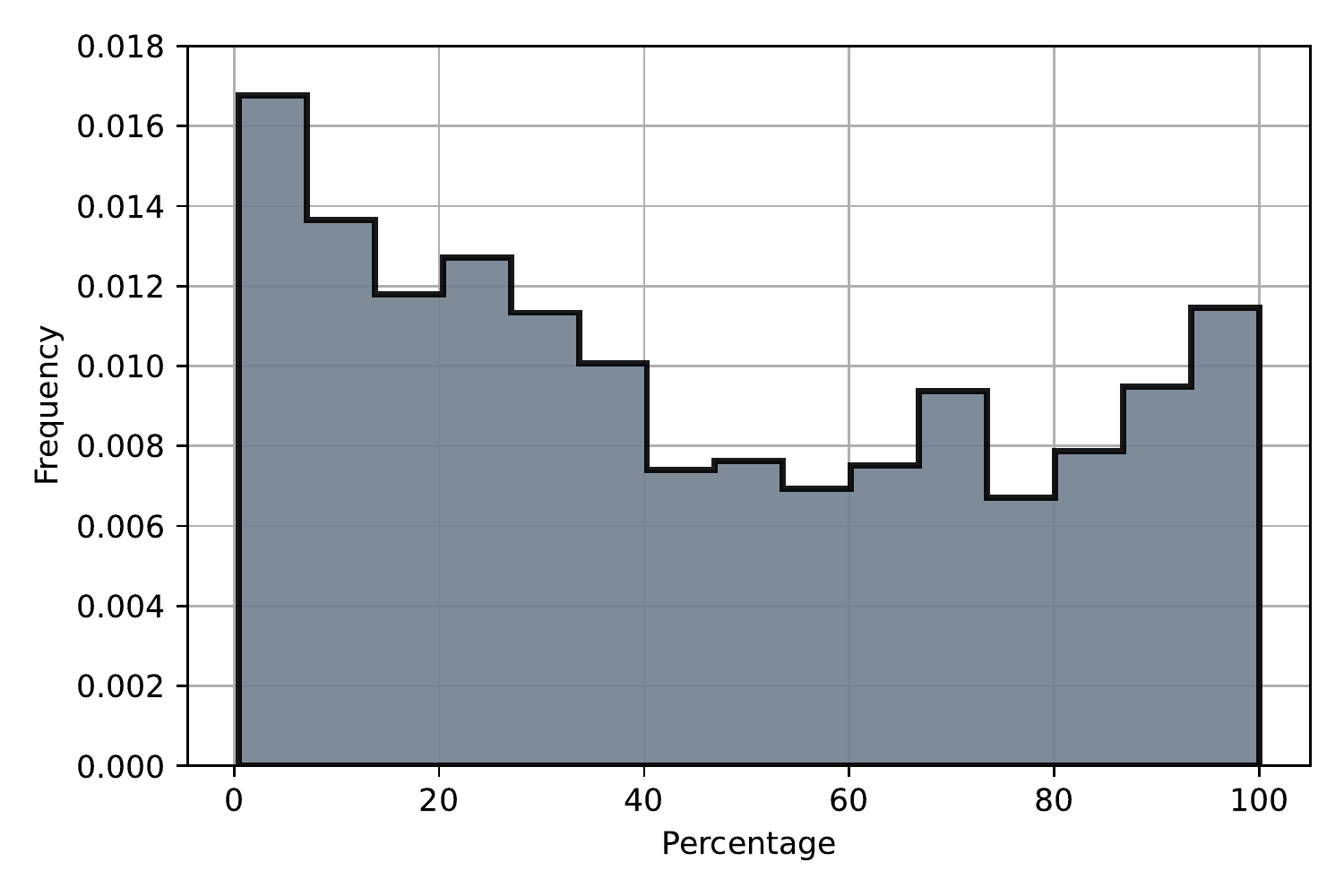}}}
    \subfloat[\textsc{Genetic}]{{\includegraphics[width=0.50\textwidth]{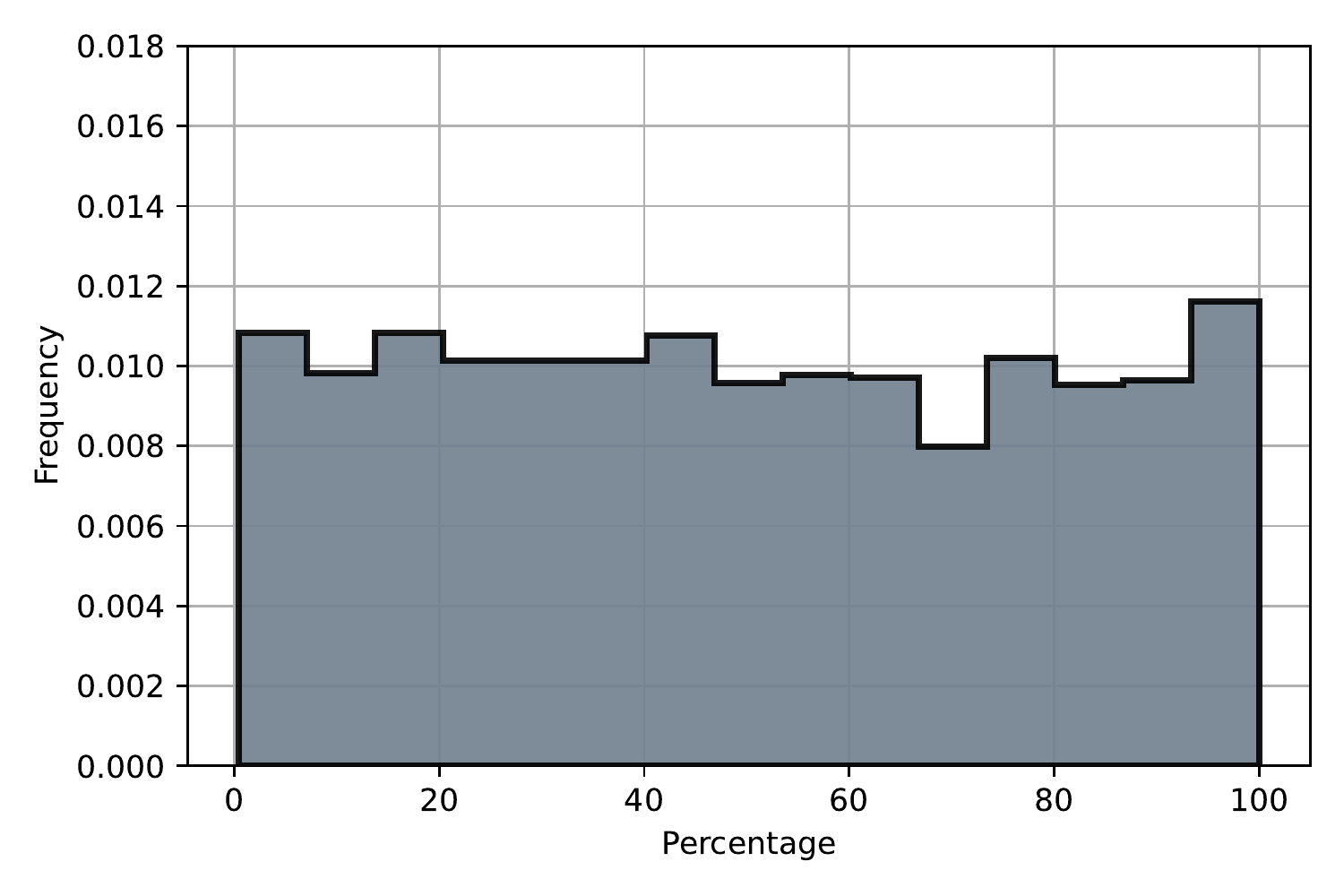}}}
}
\resizebox{1.0\textwidth}{!}{
    \subfloat[\textsc{BAE}]{{\includegraphics[width=0.29\textwidth]{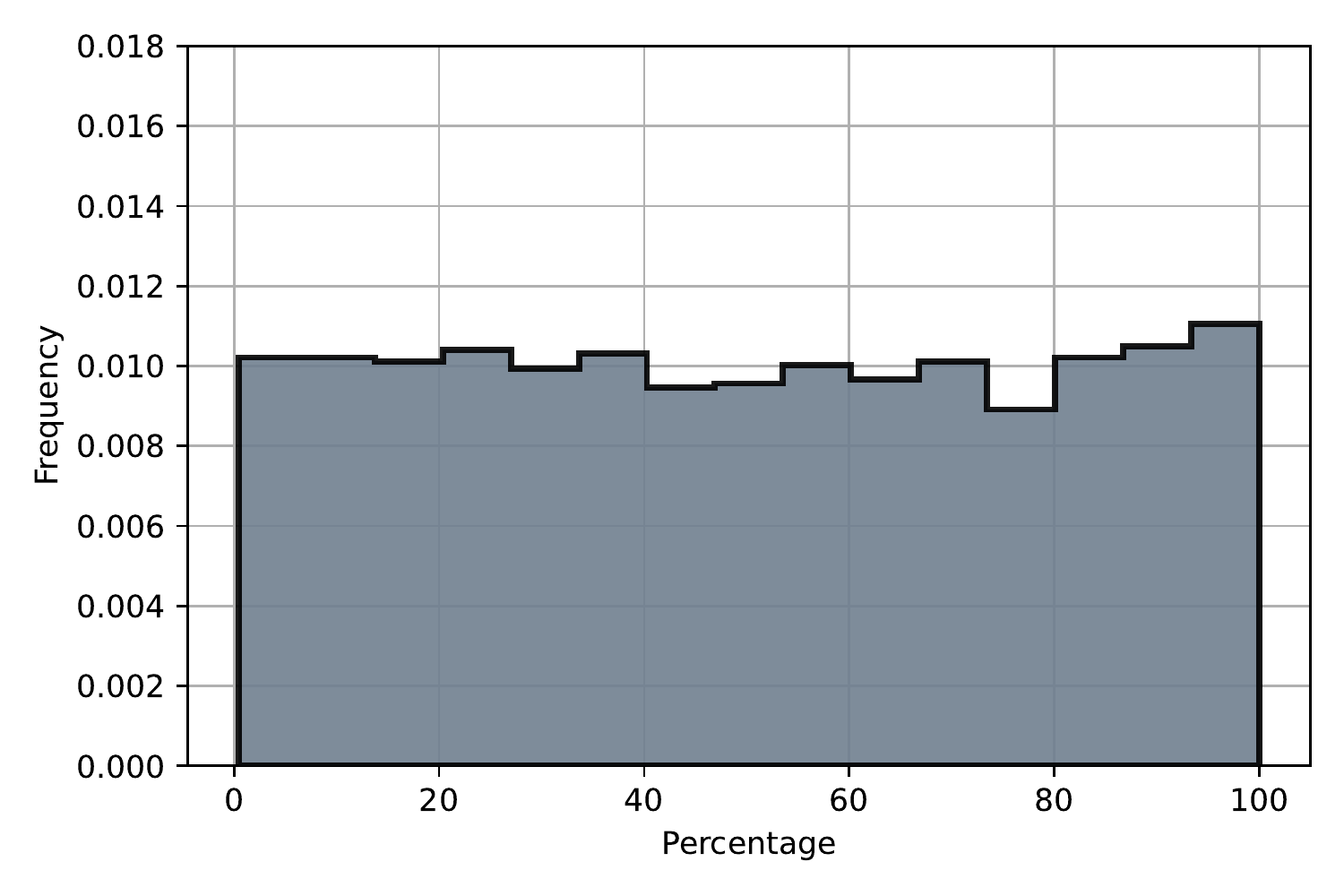}}}
    \subfloat[\textsc{TextFooler}]{{\includegraphics[width=0.29\textwidth]{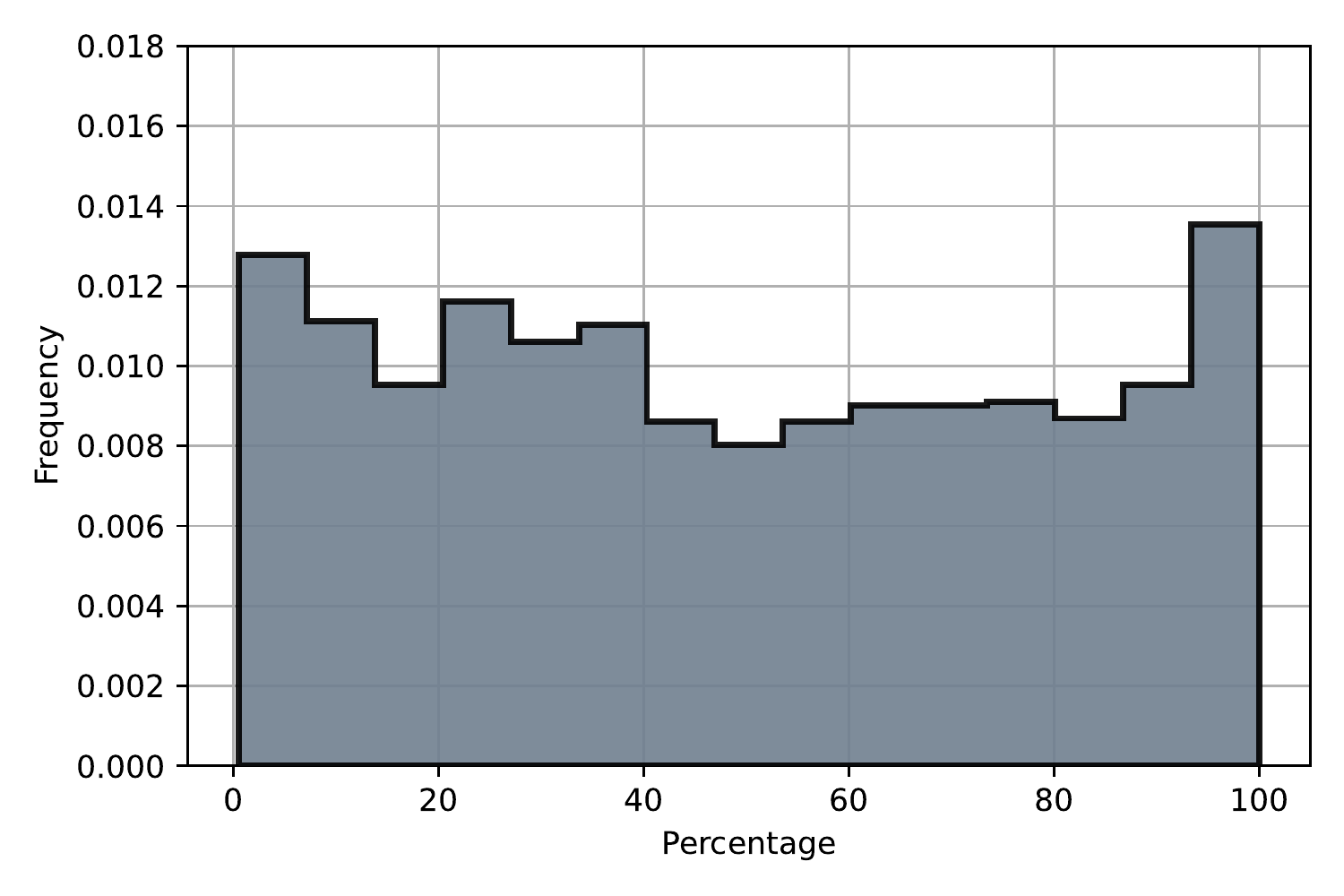}}}
    \subfloat[\textsc{SememePSO}]{{\includegraphics[width=0.29\textwidth]{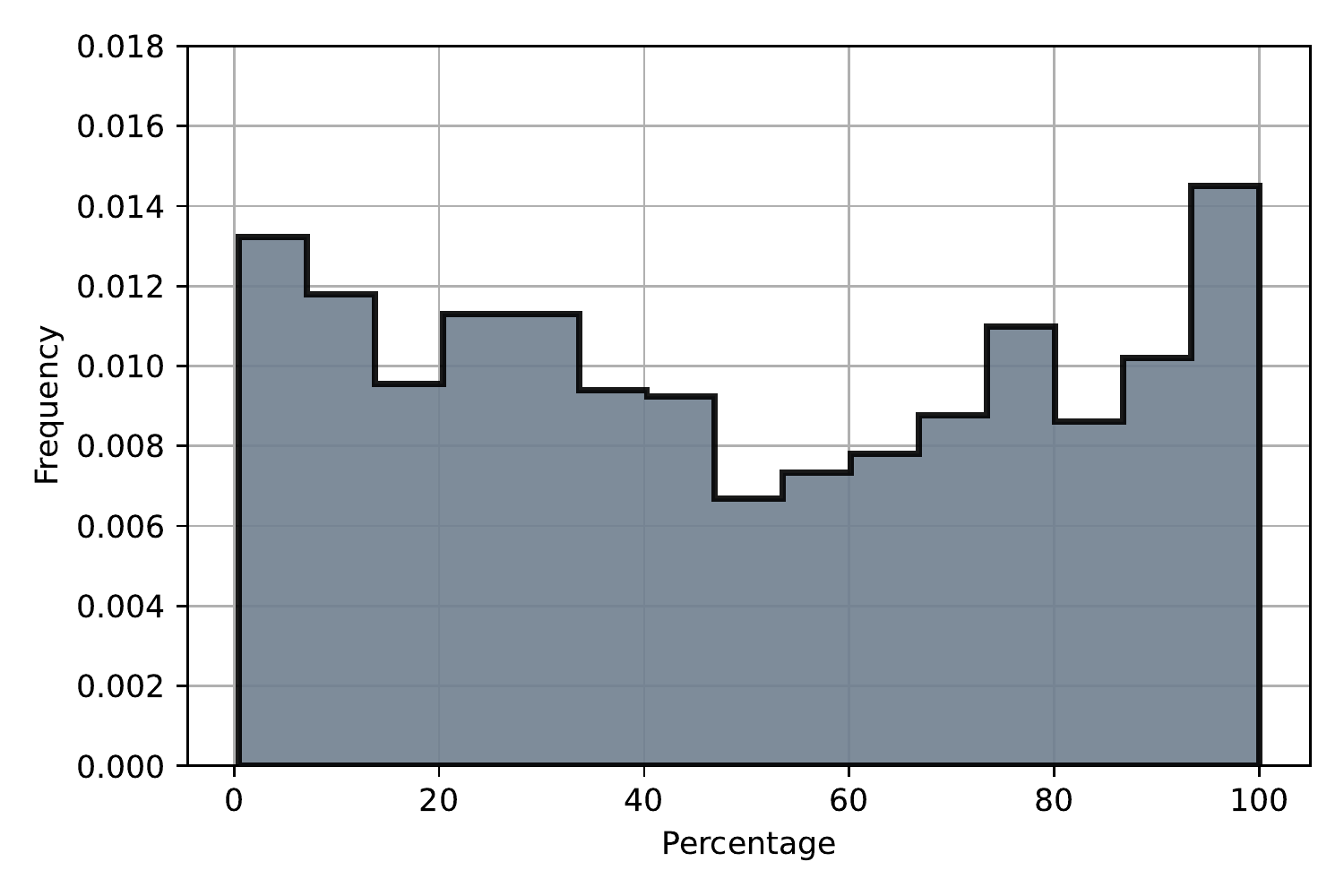}}}
}
\caption{Histograms visualising the distribution of index percentages at which the adversarial attacks perturb individual input words.}
\label{fig:hist_idx_percs}
\end{figure*}

To test this idea, we compare the pre-trained word embeddings for the replaced words and their corresponding substitutions. We choose counter-fitted \textsc{GloVe} embeddings~\cite{mrksic-etal-2016-counter}, as they push synonyms further together and antonyms further apart in representation space. 

\begin{figure*}[t]
\centering
\resizebox{0.95\textwidth}{!}{
    \subfloat[]{{\includegraphics[width=0.50\textwidth]{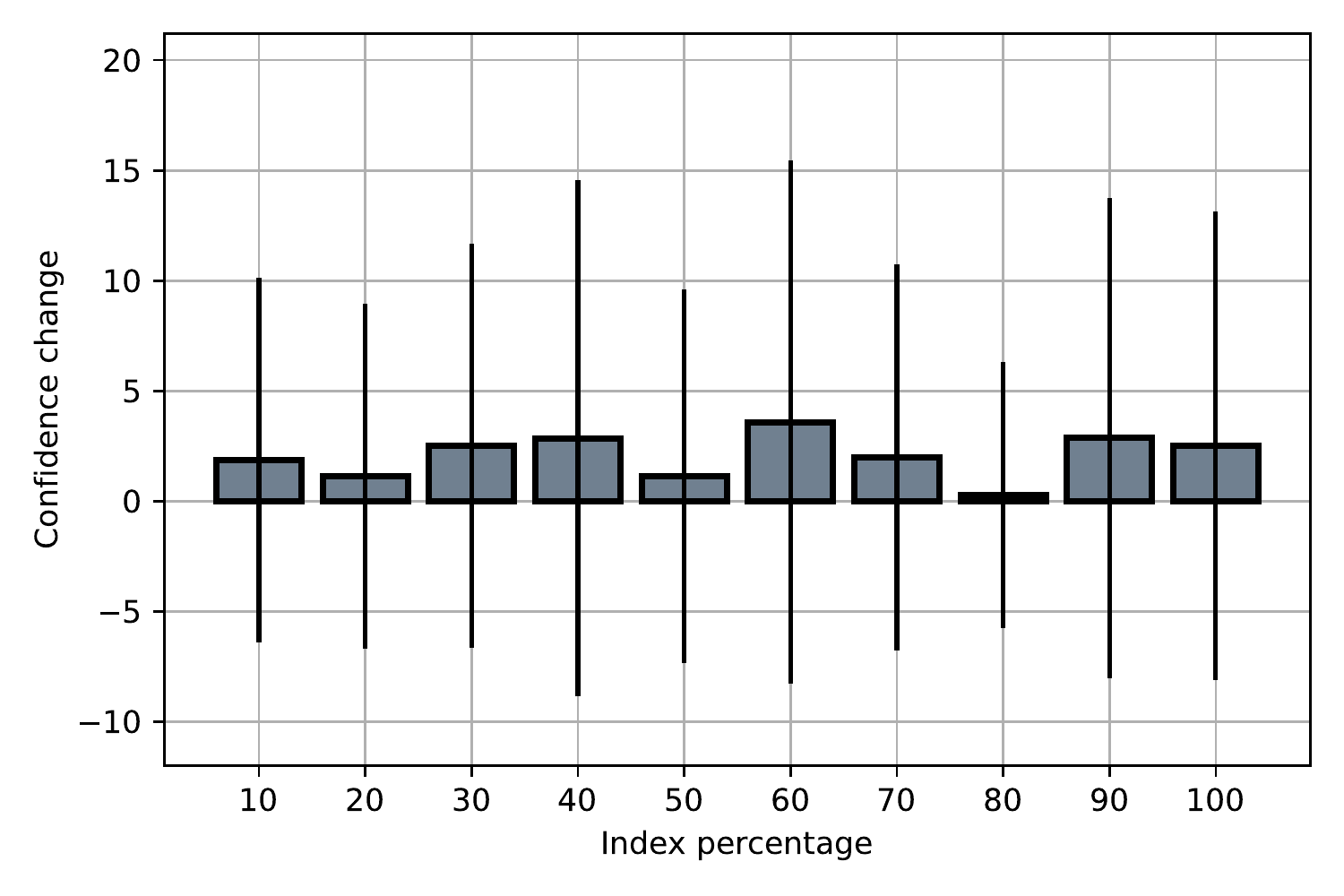}}}
    \subfloat[]{{\includegraphics[width=0.50\textwidth]{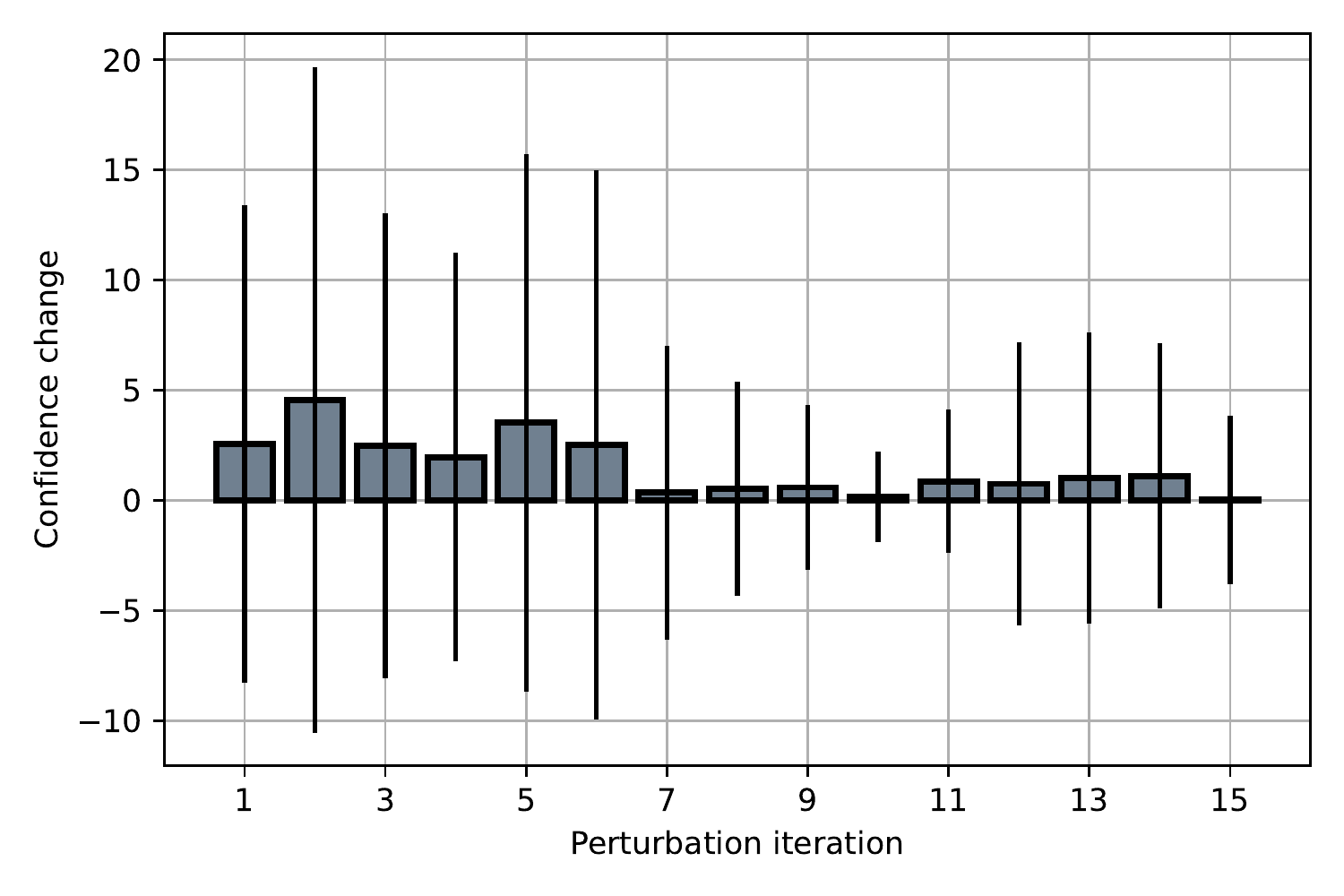}}}
}
\caption{Mean (standard deviation) prediction confidence changes on the true class across examples with respect to (a) the word index percentage and (b) the iteration in which human crowdworkers change individual input words.}
\label{fig:confidence_changes}
\end{figure*}

Table~\ref{tab:emb_dists} shows the cosine distances of the embeddings between the pairs for all five attacks. Valid pairs denotes the fraction of valid pairs used to compute the distances, since some of the word pairs did not have embedding representations in the used space. To test for statistical effects, we conduct a 5 (attacks) by 2 (success) ANOVA on the cosine distances between embeddings of replaced words and corresponding substitutions, revealing significant main effects for attack, $F(4, 6097)=363.63, p <.001$, success, $F(1, 6097)=16.43, p < .001$, as well as an interaction effect, $F(3, 6097) = 5.54, p < .001$. The entangled significant differences between attacks are indicated in Table~\ref{tab:emb_dists}. For success, a $t$-test reveals significant differences ($p<.001$) between successful and unsuccessful cosine distances across attacks. For their interaction, the difference could be driven by the lack of observations given for the unsuccessful \textsc{SememePSO} pairs.

The findings indicate that human-generated adversarial substitution pairs are significantly more similar than the substitution pairs of automated attacks (all except \textsc{Genetic}). A possible explanation for this variability is that \textsc{Genetic} uses counter-fitted embedding spaces for identifying semantically-related words for adversarial substitution. However, \textsc{TextFooler} uses the same embedding representations, yet the distances appear to be larger. Illustrative examples of semantically similar and dissimilar word substitution pairs can be found in the Appendix (Table~\ref{tab:emb_diffs}).

Repeating the analysis with regular \textsc{GloVe} embeddings yields similar results, albeit without an interaction effect (see Appendix~\ref{app:word_similarities}). We furthermore provide an analysis of sentence similarities between adversarial examples in Appendix~\ref{app:sentence_similarities}.

\subsection{How many replaced words have sentiment value?}
Particularly for the task of sentiment analysis, an attack might be more successful if it focuses on words with a sentiment value (e.g., \textit{like}, \textit{great}). We investigate the differences between attacks with respect to how many replaced words and adversarial substitutions have sentiment value. To do this, we compute the ratio of replaced words (to all replaced input words) that have a sentiment value in the NLTK sentiment lexicon~\cite{loper2002nltk}. Table~\ref{tab:sent_values} reveals that this sentiment ratio is low (between 16\% and 23\%) across attacks.

For replaced words, we observe a significant main effect for attack, $F(4, 8105)=5.28, p< .001$, but not for success or their interaction. For adversarial substitutions, the same ANOVA yields a significant effect for attack, $F(4, 8105)=54.64, p < .001$, but likewise not for success or their interaction. \textsc{HumanAdv} and \textsc{SememePSO} tend to follow that strategy more so than the remaining attacks.\footnote{This observation could potentially be explained by the finding that humans tend to over-perceive word saliencies for words with a strong sentiment value~\cite{schuff2022human}.}
We provide illustrations of the substitution pairs with the highest increases and decreases in sentiment in the Appendix (Table~\ref{tab:sent_diffs}).

\subsection{Where do humans replace?}
Next, we investigate the specific regions in an input sequence (e.g., start, middle, end) where adversarial attacks prefer to perturb words. To do this, we define the index percentage of a word in an input sequence as the ratio of the word's index to the number of words in the input (e.g., the third word of a sequence of ten words would have an index percentage of $30\%$).

Figure~\ref{fig:hist_idx_percs} shows the frequency of index percentages per attack and suggests that \textsc{HumanAdv}, \textsc{TextFooler} and \textsc{SememePSO} preferentially perturb words at the beginning and end of an input sequence. In contrast,  the distributions for \textsc{BAE} and \textsc{Genetic} show a uniform pattern. For \textsc{Genetic} this result is somewhat expected: the attack selects words for replacement by sampling words proportionately to their number of available synonyms rather than based on a semantically-informed strategy. The \textsc{HumanAdv}'s preference for replacing words at the beginning and the end of the sequence could be explained by the attention that humans devote to these parts of the text when reading from left to right. Perhaps most interestingly, the distributions for \textsc{TextFooler} and \textsc{BAE} differ, despite both using word saliencies as their word importance ranking.

We investigate which individual word changes led to notable changes in prediction confidence of the target model. We first analyse this by looking at the relationship between the index percentage and the change in prediction confidence on the true class (Figure ~\ref{fig:confidence_changes}). We observe that (a) the confidence changes caused by human perturbations are not prevalent at a specific index percentage, but rather distribute fairly evenly across the start, middle and end of the sequence. Second, Figure~\ref{fig:confidence_changes} (b) shows that the confidence changes are higher in the first iterations, and seem to drastically reduce after the sixth iteration on average. 

\section{Discussion and conclusion}
This work presented a granular analysis on strategies followed by humans when attempting to generate adversarial examples through word-level substitutions. 
We have shown that the difference in word frequency between replaced words and adversarial substitutions is smaller for humans than for the automated attacks. Furthermore, humans tend to use substitutions that are more semantically similar to the replaced words than most attacks, and humans target words that have a sentiment value to a larger extent than automated attacks.
Based on the findings provided, future directions could focus on harnessing such strategies to improve existing adversarial attacks and in doing so ultimately increase the robustness of machine learning-based NLP models against adversarial attacks.

\section*{Ethical considerations}
This paper discusses adversarial attacks in NLP, methods that are developed to uncover failure cases of machine learning models, and specifically potential approaches to further enhance such attacks against text classification models. It is worth mentioning that these methods can be used maliciously, for example to circumvent content filtering systems for hateful or offensive language on social media. Our work is intended to better understand the phenomenon of adversarial examples in NLP, its relation to human language understanding, and to harness such insights to contribute to more robust models against adversarial input perturbations.

\section*{Limitations}
The presented work comes with a number of limitations which will be discussed in this section. 

First, our analyses are limited to a single target dataset (IMDb movie reviews) and based on the only existing "human word-level adversarial attacks" dataset. Replicating our experiments on other datasets, especially those containing different styles of language use such as formal academic or journalistic writing, would help to further understand the behavioural patterns used by humans when generating adversarial examples. Future work could also build on the approach of \citet{mozes2021contrasting} to collect a larger dataset that would allow us to learn more about the strategies employed by humans when crafting adversarial examples.

Second, additional linguistic and behavioural patterns could potentially be analysed in the data. We primarily focused on the central aspects driving human strategies, yet there are other dimensions on which the data can be inspected for additional behavioural patterns (e.g., part-of-speech usage by human attackers). These are beyond the scope of this contribution but could in the future inform better attack and defence models.

Third, the dataset from ~\citet{mozes2021contrasting} did not contain potential moderating variables about the human crowdworkers. As a consequence, it is unknown how or whether differences in, for example, the language proficiency of participants, experience with NLP crowdsourcing tasks or even general cognitive abilities played a role. While the authors applied some participation requirements (i.e., participation in a similar NLP study) and trained the crowdworkers, the next step would be to understand whether psychological variables potentially moderate one's ability to craft valid adversarial examples.

Finally, the analyses in this work solely focus on statistical data analysis and do not harness data-driven machine learning-based methods to identify behavioural patterns in the data. Nevertheless, in this context the dataset size (170 human-generated sequences) represents a limitation and is potentially not large enough in size to be useful for learning-based experiments. Future work with larger datasets would mitigate that limitation and possibly help generate more insights about human strategies in adversarial example generation.

\bibliography{custom}
\bibliographystyle{acl_natbib}

\appendix

\begin{table*}[!t]
\resizebox{1.0\textwidth}{!}{
	\centering
		\begin{tabular}{c c c c c c c c c c c c c c c c}
			\toprule
		    \textbf{Difference} & 
		    \multicolumn{3}{c}{\textbf{\textsc{HumanAdv}}} &
		    \multicolumn{3}{c}{\textbf{\textsc{TextFooler}}} &
		    \multicolumn{3}{c}{\textbf{\textsc{Genetic}}} &
		    \multicolumn{3}{c}{\textbf{\textsc{BAE}}} &
		    \multicolumn{3}{c}{\textbf{\textsc{SememePSO}}} \\
			\midrule
			\multirow{ 5}{*}{High} 
			& \textit{bad} & $\to$ & \textit{,} & \textit{be} & $\to$ & \textit{sont} & \textit{one} & $\to$&  \textit{uno} & \textit{tobanga} & $\to$ & \textit{i} & \textit{movie}&  $\to$&  \textit{conga} \\
			& \textit{annoying} & $\to$ & \textit{.} & \textit{like} & $\to$ & \textit{iove} & \textit{cast} & $\to$ & \textit{foundry} & \textit{challen} & $\to$ & \textit{s} & \textit{movie} & $\to$ & \textit{cancan} \\
			& \textit{of} & $\to$ & \textit{buttery} & \textit{good} & $\to$ & \textit{buen} & \textit{action} & $\to$ & \textit{measurements} & \textit{hansika} & $\to$ & \textit{s} & \textit{really} & $\to$ & \textit{sheerly} \\
			& \textit{i} & $\to$ & \textit{i'am} & \textit{very} & $\to$ & \textit{vitally} & \textit{time} & $\to$ & \textit{timeframe} & \textit{modulates} & $\to$ & \textit{was} & \textit{film} & $\to$ & \textit{photoshoot} \\
			& \textit{this,} & $\to$ & \textit{this} & \textit{story} & $\to$ & \textit{escudos} & \textit{like} & $\to$ & \textit{adores} & \textit{bahrani} & $\to$ & \textit{t} & \textit{bad} & $\to$ & \textit{hardhearted} \\
			\midrule
			\multirow{ 5}{*}{Low} 
			& \textit{educational} & $\to$ & \textit{teaching} & \textit{frostbite} &$\to$ &\textit{frostbitten} & \textit{counselors} & $\to$ & \textit{advisors} & \textit{turns} & $\to$ & \textit{works} & \textit{appearance.the} & $\to$ & \textit{present.the} \\
			& \textit{makers} & $\to$ & \textit{producers} & \textit{movie.} & $\to$ & \textit{flick.} & \textit{wrought} & $\to$ & \textit{fabricated} & \textit{producers} & $\to$ & \textit{makers} & \textit{liked} & $\to$ & \textit{supposed} \\
			& \textit{very} & $\to$ & \textit{more} & \textit{years.i} & $\to$ & \textit{year.i} & \textit{humour} & $\to$ & \textit{mood} & \textit{low} & $\to$ & \textit{top} & \textit{manages} & $\to$ & \textit{attempts} \\
			& \textit{bad} & $\to$ & \textit{great} & \textit{rajasthan} & $\to$ & \textit{bihar} & \textit{nearly} & $\to$ & \textit{near} & \textit{match} & $\to$ & \textit{co} & \textit{promote} & $\to$ & \textit{cheer} \\
			& \textit{sing} & $\to$ & \textit{scream} & \textit{supposed} & $\to$ & \textit{felt} & \textit{dirty} & $\to$ & \textit{nasty} & \textit{dead} & $\to$ & \textit{line} & \textit{died} & $\to$ & \textit{failed} \\
			\bottomrule
		\end{tabular}
    }
	\caption{The top five pairs of replaced words and adversarial substitutions with the highest and lowest absolute frequency differences across attacks. Pairs were pre-filtered such that at least one word in a pair has a positive frequency in the training corpus, to avoid low differences due to both words having a frequency of zero.}
	\label{tab:words_freq_diffs}
\end{table*}

\begin{table*}[!t]
\resizebox{1.0\textwidth}{!}{
	\centering
		\begin{tabular}{c c c c c c c c c c c c c c c c}
			\toprule
		    \textbf{Distance} & 
		    \multicolumn{3}{c}{\textbf{\textsc{HumanAdv}}} &
		    \multicolumn{3}{c}{\textbf{\textsc{TextFooler}}} &
		    \multicolumn{3}{c}{\textbf{\textsc{Genetic}}} &
		    \multicolumn{3}{c}{\textbf{\textsc{BAE}}} &
		    \multicolumn{3}{c}{\textbf{\textsc{SememePSO}}} \\
			\midrule
			\multirow{ 5}{*}{High} 
			& \textit{in} & $\to$ & \textit{unoriginal} & \textit{like} & $\to$ & \textit{iove} & \textit{blood} & $\to$ & \textit{chrissakes} & \textit{earlier} & $\to$ & \textit{inger} & \textit{movies} & $\to$ & \textit{jitterbugs}  \\
			& \textit{adder} & $\to$ & \textit{enough} & \textit{story} & $\to$ & \textit{escudos} & \textit{brett} & $\to$ & \textit{broadly} & \textit{end} & $\to$ & \textit{oja} & \textit{box} & $\to$ & \textit{flagellation}  \\
			& \textit{back} & $\to$ & \textit{askance} & \textit{door} & $\to$ & \textit{fatma} & \textit{x} & $\to$ & \textit{tenth} & \textit{played} & $\to$ & \textit{dermott} & \textit{movie} & $\to$ & \textit{cancan}  \\
			& \textit{guard} & $\to$ & \textit{kilter} & \textit{link} & $\to$ & \textit{nol} & \textit{volunteers} & $\to$ & \textit{boneheads} & \textit{guess} & $\to$ & \textit{eses} & \textit{series} & $\to$ & \textit{wisps}  \\
			& \textit{jeepers} & $\to$ & \textit{like} & \textit{camera} & $\to$ & \textit{salas} & \textit{barbara} & $\to$ & \textit{barbaric} & \textit{put} & $\to$ & \textit{udge} & \textit{episode} & $\to$ & \textit{triviality}  \\
			\midrule
			\multirow{ 5}{*}{Low} 
			& \textit{could} & $\to$ & \textit{would} & \textit{eight} & $\to$ & \textit{six} & \textit{would} & $\to$ & \textit{could} & \textit{films} & $\to$ & \textit{film} & \textit{usually} & $\to$ & \textit{generally}  \\
			& \textit{awful} & $\to$ & \textit{terrible} & \textit{two} & $\to$ & \textit{three} & \textit{become} & $\to$ & \textit{becoming} & \textit{dancing} & $\to$ & \textit{dance} & \textit{ridiculous} & $\to$ & \textit{laughable}  \\
			& \textit{could} & $\to$ & \textit{might} & \textit{awful} & $\to$ & \textit{terrible} & \textit{awful} & $\to$ & \textit{terrible} & \textit{know} & $\to$ & \textit{tell} & \textit{positive} & $\to$ & \textit{negative}  \\
			& \textit{anything} & $\to$ & \textit{something} & \textit{test} & $\to$ & \textit{tests} & \textit{cards} & $\to$ & \textit{card} & \textit{sort} & $\to$ & \textit{kind} & \textit{specific} & $\to$ & \textit{particular}  \\
			& \textit{films} & $\to$ & \textit{film} & \textit{so} & $\to$ & \textit{too} & \textit{investment} & $\to$ & \textit{investments} & \textit{unless} & $\to$ & \textit{if} & \textit{even} & $\to$ & \textit{however}  \\
			\bottomrule
		\end{tabular}
		}
	\caption{The top five pairs of replaced words and adversarial substitutions with the highest and lowest word embedding cosine distance across attacks (using the counter-fitted embeddings).}
	\label{tab:emb_diffs}
\end{table*}

\begin{table}[!t]
\resizebox{1.0\columnwidth}{!}{
	\centering
		\begin{tabular}{l c c c c }
			\toprule
		    \textbf{Attack} & 
		    \textbf{Valid pairs} & 
		    \textbf{All} & 
		    \textbf{Succ.} &
		    \textbf{Unsucc.} \\
			\midrule
			\textsc{HumanAdv} & 1109/1303 & 0.46 (0.21) & 0.49 (0.21) & 0.45 (0.21) \\
			\textsc{TextFooler} & 1542/1805 & 0.56 (0.20) & 0.57 (0.20) & 0.52 (0.17) \\
			\textsc{Genetic} & 2020/2437 & 0.44 (0.19) & 0.45 (0.19) & 0.44 (0.19) \\
			\textsc{BAE} & 1319/1623 & 0.71 (0.30) & 0.73 (0.29) & 0.71 (0.31) \\
			\textsc{SememePSO} & 787/946 & 0.64 (0.18) & 0.64 (0.18) & -- \\
			\bottomrule
		\end{tabular}
	}
	\caption{The mean (and standard deviation) cosine distances (\textsc{GloVe} embeddings) between replaced words and corresponding substitutions for the five attacks across all perturbed sequences, divided into all, as well as successful and unsuccessful sequences.}
	\label{tab:emb_dists_glove}
\end{table}

\section{Word similarities}
\label{app:word_similarities}
We repeat the experiments in Section~\ref{sec:analysis} for word similarities with regular \textsc{GloVe} embeddings, rather than the counter-fitted ones. The mean (standard deviation) distances can be found in Table~\ref{tab:emb_dists_glove}. We here also conduct a 5 (attacks) by 2 (success) ANOVA, yielding significant effects for attack, $F(4, 6768)=371.37, p < .001$, and success, $F(1,6768)=11.27, p < .001$, but not for their interaction. To disentangle this effect for success, a subsequent test on an aggregation of successful and unsuccessful word pairs across attacks reveals significant differences ($p <.001$) between both samples. Comparing \textsc{HumanAdv} to all other attacks, we observe statistically significant ($p < .01$) differences between all comparisons for the successful portion of the data. For the unsuccessful ones, only the comparison between \textsc{HumanAdv} and \textsc{BAE} yields significant differences. 

\begin{table}[!t]
    \resizebox{1.0\columnwidth}{!}{
    	\centering
    		\begin{tabular}{l c c c }
    			\toprule
    		    \textbf{Attack} & 
    		    \textbf{All} & 
    		    \textbf{Succ.} &
    		    \textbf{Unsucc.} \\
    			\midrule
    			\textsc{HumanAdv} & 0.035 (0.050) & 0.043 (0.061) & 0.031 (0.042) \\
    			\textsc{TextFooler} & 0.064 (0.065) & 0.063 (0.064) & 0.177 (0.000) \\
    			\textsc{Genetic}$^{a}$ & 0.063 (0.052) & 0.034 (0.036) & 0.076 (0.053) \\
    			\textsc{BAE}$^{a}$ & 0.044 (0.036) & 0.022 (0.018) & 0.056 (0.039) \\
    			\textsc{SememePSO} & 0.056 (0.071) & 0.056 (0.071) & -- \\
    			\bottomrule
    		\end{tabular}
	}
	\caption{The mean (SD) cosine distances of USE representations between unperturbed and adversarial sequences. $^{a}$ indicates significant differences with \textsc{HumanAdv} for unsuccessful pairs.}
	\label{tab:sent_dists}
\end{table}

\begin{table*}[!t]
\resizebox{1.0\textwidth}{!}{
	\centering
		\begin{tabular}{c c c c c c c c c c c c c c c c}
			\toprule
		    \textbf{Sentiment increase} & 
		    \multicolumn{3}{c}{\textbf{\textsc{HumanAdv}}} &
		    \multicolumn{3}{c}{\textbf{\textsc{TextFooler}}} &
		    \multicolumn{3}{c}{\textbf{\textsc{Genetic}}} &
		    \multicolumn{3}{c}{\textbf{\textsc{BAE}}} &
		    \multicolumn{3}{c}{\textbf{\textsc{SememePSO}}} \\
			\midrule
			\multirow{ 5}{*}{Smallest} 
			& \textit{best} & $\to$ & \textit{worst} & \textit{comedic} & $\to$ & \textit{travesty} & \textit{comedy} & $\to$ & \textit{travesty} & \textit{enjoyed} & $\to$ & \textit{cut} & \textit{positive} & $\to$ & \textit{negative} \\
			& \textit{love} & $\to$ & \textit{hate} & \textit{comedy} & $\to$ & \textit{ridicule} & \textit{excited} & $\to$ & \textit{agitated} & \textit{reaches} & $\to$ & \textit{lies} & \textit{amazing} & $\to$ & \textit{horrid} \\
			& \textit{enjoyed} & $\to$ & \textit{hated} & \textit{funny} & $\to$ & \textit{odd} & \textit{intense} & $\to$ & \textit{violent} & \textit{great} & $\to$ & \textit{good} & \textit{great} & $\to$ & \textit{terrible} \\
			& \textit{excellent} & $\to$ & \textit{horrible} & \textit{comedy} & $\to$ & \textit{farce} & \textit{enlightening} & $\to$ & \textit{sobering} & \textit{brilliant} & $\to$ & \textit{worthy} & \textit{amazing} & $\to$ & \textit{terrible} \\
			& \textit{fantastic} & $\to$ & \textit{bad} & \textit{wonderful} & $\to$ & \textit{funky} & \textit{kiss} & $\to$ & \textit{screwing} & \textit{fantastic} & $\to$ & \textit{good} & \textit{wonderfully} & $\to$ & \textit{suspiciously} \\
			\midrule
			\multirow{ 5}{*}{Largest} 
			& \textit{worst} & $\to$ & \textit{best} & \textit{worst} & $\to$ & \textit{greatest} & \textit{odd} & $\to$ & \textit{curious} & \textit{bad} & $\to$ & \textit{good} & \textit{awful} & $\to$ & \textit{awesome} \\
			& \textit{bad} & $\to$ & \textit{great} & \textit{worse} & $\to$ & \textit{greatest} & \textit{strangely} & $\to$ & \textit{surprisingly} & \textit{ridiculous} & $\to$ & \textit{good} & \textit{terrible} & $\to$ & \textit{terrific} \\
			& \textit{idiotic} & $\to$ & \textit{excellent} & \textit{annoys} & $\to$ & \textit{excites} & \textit{cruel} & $\to$ & \textit{ferocious} & \textit{dead} & $\to$ & \textit{hard} & \textit{awful} & $\to$ & \textit{terrific} \\
			& \textit{poor} & $\to$ & \textit{great} & \textit{disappointments} & $\to$ & \textit{excitements} & \textit{fine} & $\to$ & \textit{beautiful} & \textit{low} & $\to$ & \textit{top} & \textit{awful} & $\to$ & \textit{thrilling} \\
			& \textit{fail} & $\to$ & \textit{excellent} & \textit{dullest} & $\to$ & \textit{neatest} & \textit{worst} & $\to$ & \textit{gravest} & \textit{worth} & $\to$ & \textit{worthy} & \textit{hard} & $\to$ & \textit{great} \\
			\bottomrule
		\end{tabular}
	}
	\caption{The top five pairs of replaced words and adversarial substitutions with the largest increases and decreases in sentiment value across attacks (based on the NLTK sentiment lexicon).}
	\label{tab:sent_diffs}
\end{table*}

\section{Sentence similarities}
\label{app:sentence_similarities}
Word similarities may only provide a limited picture as they lack context. We therefore also analyse the sentence similarity among adversarial examples. We utilise \textit{universal sentence encoder}~\cite[USE,][]{cer2018universal} representations for our analysis. Table~\ref{tab:sent_dists} shows the cosine distances for each attack type. Conducting a 5 (attack) by 2 (success) ANOVA, we observe significant effects between attacks, $F(4, 627)=6.46, p < .001$, success, $F(1, 627)=16.41, p < .001$ as well as their interaction, $F(3,627)=5.77,p < .001$.\footnote{The results of subsequent $t$-tests are indicated in Table~\ref{tab:sent_dists}.}

\end{document}